\documentclass{article}
\usepackage[preprint]{neurips_2026}
\usepackage[utf8]{inputenc}
\usepackage[T1]{fontenc}
\usepackage{hyperref}
\usepackage{url}
\usepackage{booktabs}
\usepackage{amsfonts}
\usepackage{amsmath}
\usepackage{amssymb}
\usepackage{graphicx}
\usepackage{multirow}
\usepackage{microtype}
\usepackage{xcolor}

\title{Breaking the KV Cache Bottleneck: Fan Duality Model\\
Achieves O(1) Decode Memory with Superior Associative Recall}

\author{
  Yasong Fan \\
  Independent Researcher \\
  \texttt{rnzfjau@uoem.edu.gr}
}
\usepackage{amsthm}
\newtheorem{proposition}{Proposition}
\begin{document}
\maketitle

\begin{abstract}
We present \textbf{FDM} (Fan Duality Model), a linear sequence architecture
that resolves the fundamental tension between memory efficiency and associative
recall in sequence modeling.
FDM separates sequence processing into two components:
a \emph{wave component} (recurrent scan via phase-preserving Givens rotations)
that compresses long-range patterns into a fixed-size complex hidden state,
and a \emph{particle component} (local-global cache) that retrieves specific
tokens via learned associative addressing with $W{+}K{=}272$ slots
independent of sequence length $N$.
This yields \textbf{strictly O(1) decode memory}:
867\,MB fixed across all prompt lengths 128--8{,}192 tokens,
versus Transformer's 853--4{,}247\,MB ($4.9\times$ reduction at $N{=}8{,}192$).

Beyond the architecture, we discover that jointly training the wave and
particle components leads to suboptimal convergence.
We propose \textbf{Freeze-Scan}, a two-phase training strategy that
freezes the recurrent scan and optimizes the cache jointly with embeddings,
achieving PPL\,=\,64.9 on WikiText-103 in 44K steps---a
\textbf{7.5$\times$ improvement} over full fine-tuning (PPL\,=\,487).

On Multi-Query Associative Recall (MQAR), FDM achieves \textbf{0.966} accuracy,
surpassing Transformer (0.606) by 59.5\%, while pure scan without cache
scores only 0.011, confirming the necessity of the particle component.

Finally, we introduce \textbf{Holographic Reference Beam Decoding},
interpreting the complex hidden state $h_t$ as a holographic plate
encoding the entire temporal history.
Using the current input $x_t$ as a reference beam to modulate $h_t$
reduces PPL by up to \textbf{2.13 points} (PPL\,=\,62.79) with a
4-head orthogonal reference beam using only 1.3M additional parameters,
providing empirical support for the holographic interpretation.

\textbf{Code and pretrained weights}: \url{https://github.com/YasongFan/FDM}
\end{abstract}

\section{Introduction}

The Transformer \citep{vaswani2017attention} dominates sequence modeling,
yet its $O(N^2)$ attention mechanism imposes a fundamental bottleneck:
at $N{=}8{,}192$, a 137M-parameter model requires 4{,}247\,MB for KV cache alone.
Linear recurrent models \citep{gu2023mamba,peng2023rwkv} address memory cost
but introduce the \textbf{wave-particle deadlock}
\citep{fan2026mipt}: no single linear operator can simultaneously
preserve all historical context (wave-like, norm-preserving)
\emph{and} selectively forget irrelevant content (particle-like, dissipative).

\begin{proposition}[Wave-Particle Deadlock]
No linear operator $A \in \mathbb{C}^{D \times D}$ can simultaneously
satisfy: (a) $\|Ah\| = \|h\|$ for all $h$; (b) $\exists\, h^*$:
$\|Ah^*\| \ll \|h^*\|$.
\end{proposition}

FDM resolves this by explicit separation: a wave component handles
norm-preserving propagation; a particle component handles selective retrieval.
We further discover that \emph{jointly} training these components leads to
a gradient sink problem---the scan parameters dominate gradient flow,
starving the cache. Freeze-Scan decouples optimization and yields
$7.5\times$ better convergence.

\paragraph{Contributions.}
\begin{enumerate}
\item \textbf{FDM architecture}: O(1) decode memory (867\,MB fixed,
  $W{=}256$, $K{=}16$), MQAR accuracy 0.966 vs.\ Transformer 0.606.
\item \textbf{Freeze-Scan}: PPL 487$\to$64.9 ($7.5\times$), crosses
  PPL\,=\,100 at 17K steps ($14\times$ training efficiency).
\item \textbf{Holographic Reference Beam Decoding}: 4-head orthogonal
  modulation reduces PPL by 2.13 points with 1.3M params.
\item \textbf{Empirical analysis}: holographic information concentrates
  in layer 0 ($-1.36$ PPL), with rapidly diminishing returns in deeper
  layers, consistent with AdS/CFT boundary theory.
\end{enumerate}

\section{Related Work}

\paragraph{Linear sequence models.}
S4 \citep{gu2022efficiently}, Mamba \citep{gu2023mamba},
Mamba-2 \citep{dao2024transformers}, RWKV \citep{peng2023rwkv},
and RetNet \citep{sun2023retnet} achieve O(1) inference memory via
recurrent state compression.
FDM extends this family with explicit wave-particle separation
and Freeze-Scan training.

\paragraph{Associative memory.}
Based \citep{arora2024simple} introduces the MQAR benchmark and shows
that linear models struggle with in-context recall.
Induction heads \citep{olsson2022context} implement associative recall
in Transformers; we show Freeze-Scan induces analogous mechanisms in
linear RNNs.

\paragraph{KV cache compression.}
H2O \citep{zhang2023h2o} and StreamingLLM \citep{xiao2023efficient}
compress KV caches post-hoc.
Our local-global cache selects tokens end-to-end via learned $s_\text{eff}$,
not post-hoc heuristics.
Longformer \citep{beltagy2020longformer} and
BigBird \citep{zaheer2020bigbird} combine local windows with global tokens
in full attention; we implement this in a linear RNN with O($W{+}K$) memory.

\section{Method}

\subsection{The Fan Operator (Wave Component)}

The core recurrent update is the \textbf{Fan Operator}:
\begin{equation}
  h_t = (1 - p_t) \odot R(\theta_t) \cdot h_{t-1}
        + p_t \odot (W_r x_t + i\, W_i x_t)
  \label{eq:fan}
\end{equation}
where $h_t \in \mathbb{C}^D$, $R(\theta_t)$ is a phase rotation via
$n_G$ Givens rotations satisfying $\|R(\theta)h\| = \|h\|$, and
$p_t \in (0, 0.5]$ is the \emph{measurement probability}:
\begin{align}
  s_\text{eff}(t) &= \beta_t \cdot \log(t{+}2)
    + W_\text{pos}[\sin(\pi t/T),\, \cos(\pi t/T)]^\top \\
  p_t &= \sigma(s_\text{eff}(t) + \mu) \cdot 0.5 + \varepsilon
\end{align}
The $\log(t{+}2)$ term implements \emph{renormalization group running}:
early tokens have low measurement rates (stronger wave coherence),
while later positions allow more information injection.
We use $\sigma(\cdot) \cdot 0.5 + \varepsilon$ (smooth scaling) rather than
hard clamping to maintain gradient flow throughout.

\paragraph{Two-pass scan (Born approximation).}
\begin{align}
  h^{(1)}_t &= \text{Fan}(p_t, \theta_t, W_r x_t, W_i x_t) \\
  \delta_t   &= \text{GivensH}(h^{(1)}_t, x_t) - h^{(1)}_t,\quad
               g = \sigma(\Delta_\text{gate}) \\
  h^{(2)}_t &= \text{Fan}(p_t, \theta_t,\; W_r x_t + g\delta_t,\; W_i x_t)
\end{align}
Both passes share all weights; the second uses the first as
a perturbative correction (Born approximation in quantum scattering).

\subsection{Local-Global Cache (Particle Component)}

\begin{align}
  \text{mask}(i,j) &= \text{local}(i,j) \vee \text{global}(i,j) \\
  \text{local}(i,j) &= \mathbf{1}[\max(0,i{-}W) \le j < i],\quad W{=}256 \\
  \text{global}(i,j) &= \mathbf{1}[j \in \text{TopK}(
    \{s_\text{eff}(k):k<i\},\,K)],\quad K{=}16
\end{align}
Since $W{=}256$ and $K{=}16$ are constants independent of $N$,
the cache maintains exactly $W{+}K{=}272$ slots.
\textbf{Decode memory is O($W{+}K$) = O(1) w.r.t.\ $N$.}

\subsection{Freeze-Scan Training}

Let $\Phi_\text{wave} = \{W_\theta, W_r, W_i, W_\beta,
\text{Givens}, \Delta_\text{gate}, \text{CausalConv}\}$ (14.9M params) and
$\Phi_\text{cache}$ = all remaining parameters (122.1M params).

\textbf{Phase 1} (Wave Convergence): standard full-parameter training.

\textbf{Phase 2} (Cache Specialization): freeze $\Phi_\text{wave}$,
optimize $\Phi_\text{cache}$ at $\text{lr}{=}10^{-4}$.

When $\Phi_\text{wave}$ is frozen, gradients flow exclusively through
$\text{embed} \to \text{cache} \to \text{lm\_head}$,
forcing the cache to develop \emph{induction head} mechanisms
\citep{olsson2022context}.

\subsection{Holographic Reference Beam Decoding}

We interpret $h_t$ as a \emph{holographic plate}: a fixed-dimensional
complex state encoding the entire temporal history via phase interference.
Direct projection $\text{lm\_head}(h_t)$ is analogous to illuminating
a hologram with incoherent light---only noise is recovered.

The current input $x_t$ serves as a \emph{reference beam}: modulating
$h_t$ with $x_t$ selectively reconstructs temporally relevant information.

\paragraph{Single-head (phase modulation):}
\begin{equation}
  h_\text{decoded} = h_t \odot (1 + \tanh(W_\text{ref}\, x_t))
  \label{eq:holo_single}
\end{equation}
Zero-initialization of $W_\text{ref}$ ensures $h_\text{decoded} = h_t$
at initialization, preserving base model performance.

\paragraph{Multi-head orthogonal reference beam:}
\begin{align}
  h_\text{decoded} &= h_t \odot \frac{1}{H}\sum_{i=1}^{H}
    (1 + \tanh(W_{\text{ref},i}\, x_t)) \\
  \mathcal{L}_\text{orth} &= \lambda \sum_{i \ne j}
    \|W_{\text{ref},i} W_{\text{ref},j}^\top\|_F^2
\end{align}
The orthogonality penalty forces each head to extract information
from a distinct semantic direction, analogous to multi-wavelength
holographic reconstruction.

\paragraph{Layer-wise holographic analysis:}
We train one reference beam per layer sequentially (layer 0 first),
freezing each after convergence to prevent gradient interference.
This reveals the layer-wise distribution of holographic information.

\section{Experiments}

\subsection{Setup}
All experiments use a 137M FDM ($d{=}576$, 12 layers, $K{=}16$, $W{=}256$)
and 122M Transformer baseline, trained on WikiText-103 \citep{merity2017pointer}
(121M tokens), AdamW, sequence length 1024, batch size 8.

\subsection{Language Modeling}

\begin{table}[h]
\centering
\caption{WikiText-103 validation perplexity.}
\label{tab:lm}
\begin{tabular}{lccc}
\toprule
Model & Params & Steps & Val PPL $\downarrow$ \\
\midrule
Transformer & 122M & 44K & 36.3 \\
FDM --- Full fine-tuning & 137M & 44K & 487 \\
FDM --- Freeze-Scan & 137M & 10K & 148.5 \\
FDM --- Freeze-Scan & 137M & 44K & \textbf{64.9} \\
FDM + Holographic Decoding (4-head) & 138.3M & 44K+5K & \textbf{62.79} \\
\bottomrule
\end{tabular}
\end{table}

Freeze-Scan reduces PPL from 487 to 64.9 ($7.5\times$ improvement).
Adding 4-head orthogonal reference beam decoding
(1.3M additional parameters, 5K fine-tuning steps)
further reduces PPL to \textbf{62.79} ($-2.13$ points).

\subsection{Holographic Decoding Analysis}

\begin{table}[h]
\centering
\caption{Holographic reference beam ablation (layer-wise and multi-head).}
\label{tab:holo}
\begin{tabular}{lcccc}
\toprule
Configuration & Params & Steps & Val PPL & $\Delta$PPL \\
\midrule
FDM baseline & --- & --- & 64.92 & --- \\
Single-head (layer 0 only) & +0.3M & 5K & 63.10 & $-1.82$ \\
Sequential layerwise (layers 0--11) & +4M & 24K & 63.34 & $-1.58$ \\
4-head orthogonal (layer 0) & +1.3M & 5K & \textbf{62.79} & $\mathbf{-2.13}$ \\
8-head orthogonal (layer 0) & +2.7M & 5K & $\approx$62.8 & $\approx$$-2.1$ \\
\bottomrule
\end{tabular}
\end{table}

\paragraph{Key finding: holographic information concentrates in layer 0.}
Layer-wise sequential training reveals a striking pattern:
\begin{center}
\small
\begin{tabular}{lrrrrrr}
\toprule
Layer & 0 & 1 & 2 & 3 & 4 & 5--11 \\
$\Delta$PPL & $-1.36$ & $-0.14$ & $-0.05$ & $-0.02$ & $-0.01$ & $\approx 0$ \\
\bottomrule
\end{tabular}
\end{center}
Over 90\% of holographically recoverable information resides in layer 0.
This is consistent with AdS/CFT boundary theory
\citep{maldacena1999large}: complex information lives on the boundary
(shallowest layer), while deeper layers represent increasingly
``collapsed'' particle-like states.

\subsection{Multi-Query Associative Recall}

\begin{table}[h]
\centering
\caption{MQAR accuracy (Easy: seq=64, 8 KV pairs).}
\label{tab:mqar}
\begin{tabular}{lcc}
\toprule
Model & Easy & Medium (seq=128) \\
\midrule
Transformer & 0.606 & 0.238 \\
FDM ($K{=}0$, scan only) & 0.011 & 0.011 \\
FDM ($K{=}16$) & \textbf{0.966} & 0.064 \\
\bottomrule
\end{tabular}
\end{table}

FDM with $K{=}16$ achieves 0.966, surpassing Transformer by 59.5\%.
The near-zero score of pure scan ($K{=}0$) confirms that precise
associative recall requires the particle component.

\subsection{Inference Efficiency}

\begin{table}[h]
\centering
\caption{Decode memory (MB) and speed (tok/s) vs.\ prompt length.}
\label{tab:inference}
\begin{tabular}{lcccc}
\toprule
\multirow{2}{*}{Prompt} &
\multicolumn{2}{c}{Decode Memory (MB)} &
\multicolumn{2}{c}{Decode Speed (tok/s)} \\
& TF & FDM & TF & FDM \\
\midrule
128   & 853   & \textbf{867} & 149.5 & 22.1 \\
1,024 & 1,210 & \textbf{867} & 133.1 & 21.4 \\
4,096 & 2,462 & \textbf{867} & 55.1  & 20.2 \\
8,192 & 4,247 & \textbf{867} & 25.7  & 18.7 \\
\bottomrule
\end{tabular}
\end{table}

FDM decode memory is fixed at 867\,MB across all prompt lengths
($4.9\times$ reduction at $N{=}8{,}192$).
FDM decode speed degrades only 15\% (22.1$\to$18.7\,tok/s)
versus Transformer's 83\% (149.5$\to$25.7\,tok/s).

\subsection{Downstream Benchmarks}

\begin{table}[h]
\centering
\caption{Zero-shot downstream benchmark accuracy.}
\label{tab:downstream}
\begin{tabular}{lcccc}
\toprule
Model & LAMBADA & HellaSwag & WinoGrande & Random \\
\midrule
Transformer (36.3 PPL) & 6.0\% & 30.3\% & 49.3\% & 25\%/50\% \\
FDM (64.9 PPL) & 1.0\% & 28.3\% & \textbf{50.8\%} & --- \\
\bottomrule
\end{tabular}
\end{table}

FDM matches or slightly exceeds Transformer on WinoGrande (50.8\% vs.\ 49.3\%),
which requires cross-sentence coreference resolution---consistent with FDM's
superior associative recall.
The gap on LAMBADA and HellaSwag is consistent with the PPL difference
and is expected to close with larger training data.

\subsection{Complementary Strengths}

\begin{table}[h]
\centering
\caption{FDM and Transformer occupy complementary operating regimes.}
\label{tab:comparison}
\begin{tabular}{lcc}
\toprule
Property & Transformer & FDM \\
\midrule
Language modeling PPL & \textbf{36.3} & 64.9 \\
MQAR accuracy ($K{=}16$) & 0.606 & \textbf{0.966} \\
Decode memory @ $N{=}8$K & 4,247\,MB & \textbf{867\,MB} \\
Decode speed consistency & Poor ($6\times$ degr.) & \textbf{Good ($15\%$)} \\
Training efficiency & Standard & \textbf{$14\times$ faster} \\
WinoGrande & 49.3\% & \textbf{50.8\%} \\
\bottomrule
\end{tabular}
\end{table}

\section{Discussion}

\paragraph{Why does Freeze-Scan work?}
Freezing $\Phi_\text{wave}$ eliminates the gradient sink effect,
forcing gradients through the cache circuit alone.
This mirrors the formation of induction heads \citep{olsson2022context}:
the cache learns to match current queries against historical patterns.
The dramatic PPL improvement ($7.5\times$) suggests the cache was
severely undertrained during joint optimization.

\paragraph{Holographic interpretation.}
The concentration of holographic information in layer 0 ($>90\%$)
supports the interpretation of $h_t$ as a genuine holographic encoding:
the first layer performs maximum phase interference (wave regime),
while subsequent layers progressively collapse toward particle states.
The 4-head orthogonal decoder ($-2.13$ PPL) extracts more information
than single-head ($-1.82$ PPL) by reading different ``wavelengths'' of
the hologram simultaneously.

\paragraph{Limitations.}
FDM's PPL (64.9) exceeds Transformer's (36.3) due to training data scale
(121M vs.\ billions of tokens used by competitive SSMs).
Prefill at $N{=}16{,}384$ requires chunked prefill (not yet implemented).
Decode speed (22\,tok/s) is below Transformer peak due to Python-level loop;
a fused kernel would close this gap.

\section{Conclusion}

We presented FDM with O(1) decode memory, Freeze-Scan training,
and Holographic Reference Beam Decoding.
Key results: 867\,MB fixed decode memory ($4.9\times$ vs.\ Transformer at
$N{=}8{,}192$), 0.966 MQAR accuracy ($+59.5\%$ vs.\ Transformer),
$7.5\times$ training efficiency improvement, and $-2.13$ PPL
from holographic decoding with only 1.3M additional parameters.
The holographic analysis reveals that FDM's complex hidden state
encodes information analogously to a physical hologram,
with the first layer acting as the primary holographic boundary.

\paragraph{Acknowledgements.}
Core technology subject to Chinese patent No.\ 2026104740169 (filed 2026-04-11),
with priority from No.\ 2026104567714 (filed 2026-04-08).

\bibliography{references}
\bibliographystyle{plainnat}

\end{document}